\documentclass[10pt,twocolumn,letterpaper]{article}

\usepackage{cvpr}              % To produce the CAMERA-READY version
\usepackage{suffix}
\usepackage{etoolbox}
\definecolor{cvprblue}{rgb}{0.21,0.49,0.74}
\usepackage[pagebackref,breaklinks,colorlinks,allcolors=cvprblue]{hyperref}

\usepackage{pifont}% http://ctan.org/pkg/pifont
\newcommand{\xmark}{\ding{55}}%
\usepackage{algorithm}
\usepackage{algpseudocode}
\usepackage{multirow}
\usepackage{adjustbox}
\usepackage{array}
\usepackage{comment}
\usepackage{microtype}
\usepackage{bbm}
\newcolumntype{P}[1]{>{\raggedright\arraybackslash}p{#1}}
\usepackage{setspace}
\usepackage[accsupp]{axessibility}

\usepackage[dvipsnames]{xcolor}

\def\paperID{65} % *** Enter the Paper ID here
\def\confName{CVPR}
\def\confYear{2026}

\title{AOI-SSL: Self-Supervised Framework for Efficient Segmentation of Wire-bonded Semiconductors In Optical Inspection}

\author{
    Joaquín Figueira$^1$ \quad 
    Rob van Gastel$^2$ \quad 
    Giacomo D'Amicantonio$^1$ \quad 
    Zhuoran Liu$^3$ \\
    Ioan Gabriel Bucur$^3$ \quad 
    Faysal Boughorbel$^2$ \quad 
    Egor Bondarev$^1$ \\[2pt]
    $^1$AIMS Lab, Eindhoven University of Technology \quad $^2$CoC, ASMPT \quad $^3$iCIS, Radboud University \\
    {\tt\small \{j.figueira, g.d.amicantonio, e.bondarev\}@tue.nl} \\
    {\tt\small \{rob.van.gastel, faysal.boughorbel\}@asmpt.com} \quad
    {\tt\small \{zhuoran.liu, gabriel.bucur\}@ru.nl}
    \vspace{-10pt}
}
\begin{document}
\maketitle
\begin{abstract}
Segmentation models in automated optical inspection of wire-bonded semiconductors are typically device-specific and must be re-trained when new devices or distribution shifts appear. We introduce AOI-SSL, a training-efficient framework for semantic segmentation of wire-bonded semiconductors by combining small-domain self-supervised pre-training of vision transformers with in-context inference that minimizes the need of labeled examples. We pre-train SOTA self-supervised algorithms in a small industrial inspection dataset and find that Masked Autoencoders are the most effective in this small-data setting, improving downstream segmentation while reducing the labeled fine-tuning effort. We further introduce in-context, patch-level retrieval methods that predict masks directly from dense encoder embeddings with negligible additional training. We show that, in this setting, simple similarity-based retrieval performs on par with more complex attention-based aggregation used currently in the literature. Furthermore, our experiments demonstrate that self-supervised pre-training significantly improves segmentation quality compared to training from scratch and to ImageNet pre-trained backbones under a fixed fine-tuning computational budget. Finally, the results reveal that retrieval based segmentation outperforms fine-tuning when targeting single device images, allowing for near-instant adaptation to difficult samples. Code is available at \url{https://github.com/jacomof/aoi-ssl}.
\end{abstract}    
\section{Introduction}

Automated optical inspection (AOI) is a critical quality control component in semiconductor manufacturing, where high-resolution images are used to detect defects and structural anomalies during and after packaging. In the context of wire-bonded devices, semantic segmentation of components such as wires, bonds, wedges, and encapsulation material is often a prerequisite for downstream inspection and decision-making. 
Despite recent progress in deep learning-based inspection systems, most AOI pipelines still rely on models and preprocessing routines that are tightly coupled to specific device designs and acquisition conditions \cite{Zhang2024, Chen2020, Wu2020, Cai2018}. As a consequence, introducing a new device or encountering a significant covariate shift typically requires collecting new labeled data and re-training new segmentation models, which is costly and time-consuming in production environments.

\begin{figure}[t]
    \centering
    \includegraphics[width=1.0\linewidth]{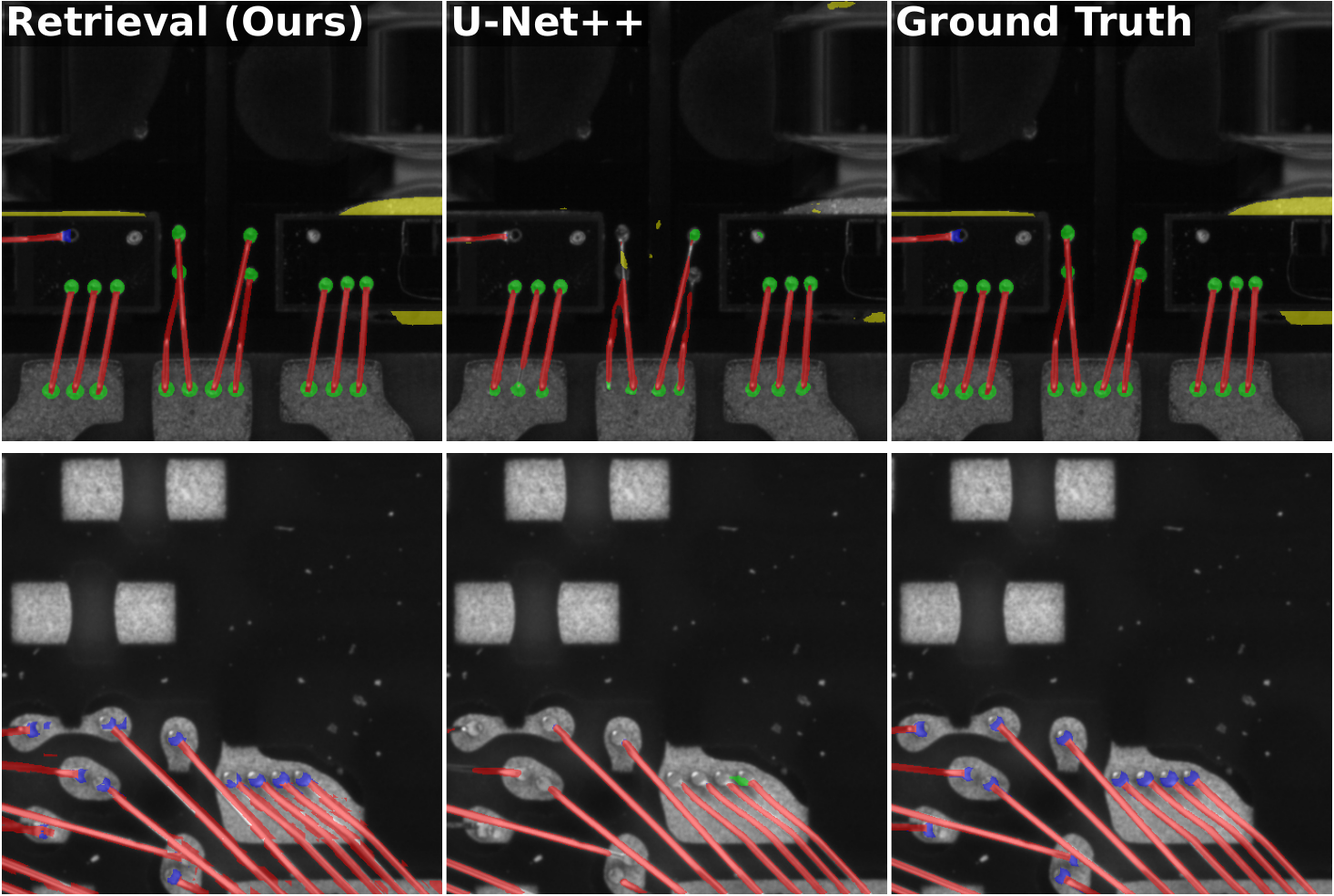}
    \caption{\textbf{Segmentation Performance on Complex Wire-bond Geometry}. Different classes are highlighted in different colors over monochrome images of representative samples. Our retrieval method shows superior performance in two difficult devices where the baseline ResNet18 + UNet++ model fails to segment \textbf{\textcolor{blue}{wedge bonds}} entirely.}
    \label{fig:paper-teaser}
    \vspace{-1em}
\end{figure}

Recent advances in self-supervised learning (SSL) have demonstrated that strong visual representations can be learned without manual annotation by pre-training on large-scale image collections~\cite{Chen2020simple, he2020momentum, grill2020bootstrap, simeoni2025dinov3}. Methods based on self-distillation and Masked Image Modeling (MIM), such as DINOv2~\cite{oquab2023dinov2}, Masked Autoencoders (MAE)~\cite{He2022MAE}, and iBOT~\cite{Zhou2022ibotimagebertpretraining}, have shown impressive transfer performance in a variety of downstream tasks.
However, most existing work assumes access to very large and diverse pre-training datasets~\cite{gui2024survey}, or the existence of pre-trained models adequate in their image modality (usually RGB images). In contrast, AOI systems in the semiconductor industry typically operate in highly specialized visual domains, where only a limited amount of labeled and unlabeled data is available~\cite{nvidiaOptimizingSemiconductor}, and where imaging data does not follow the standard RGB structure (e.g., X-Ray imaging~\cite{Chen2020}). In the context of industrial optical inspection, monochrome cameras are widely used because they provide superior sensitivity, resolution, and processing speed~\cite{vacho2018selected}, adding an additional differentiator with the most common computer vision algorithms.

To tackle these challenges, we introduce AOI-SSL, a comprehensive framework tailored for domain-specific automated optical inspection. Our approach integrates self-supervised pre-training and fine-tuning strategies designed for low-data regimes, utilizing efficient Vision Transformer (ViT)~\cite{Dosovitskiy2021} architectures and in-context test-time adaptation. Specifically, we investigate whether domain-specific SSL can yield representations that rival or surpass those of established encoders pre-trained on massive, generic datasets, such as ImageNet~\cite{ImageNet}. This is evaluated in the context of semantic segmentation for wire-bonded semiconductor devices using a non-standard, monochrome 2-channel format. In this configuration, each channel captures a grayscale view of the device under a distinct light source, a setup optimized for optical calibration and depth estimation in the broader inspection pipeline. \cref{fig:paper-teaser} illustrates these dual-light views along with their corresponding semantic segmentation masks for two representative devices.

Our focus is on improving the downstream performance of the encoders for semantic segmentation and minimizing the amount of costly human-labeled data. We show that, in this setting, MAE pre-training is particularly effective, leading to faster convergence and higher segmentation performance compared to alternative pre-training strategies. 

Beyond parametric fine-tuning, retrieval-based segmentation strategies, which leverage the encoder representation space directly at inference time, are explored for fast test-time adaptation. Inspired by recent work on retrieval-based scene understanding~\cite{Balazevic2023towards, pariza2025nearfarpatchorderingenhances}, we propose patch-level retrieval methods that operate on dense token embeddings and produce segmentation masks by aggregating labels from nearest neighbors in an embedding key-value memory bank.
Unlike conventional encoder--decoder architectures, these approaches require little to no additional training on labeled data and can be adapted to new devices with only a relatively small number of labeled images by updating the retrieval memory bank.

Finally, we analyze architectural inductive bias in this domain and show that transformer variants that incorporate convolutional elements, such as FasterViT~\cite{Hatamizadeh2023}, yield significant gains for fine-grained structures over traditional vision transformers. Our contributions can be summarized as follows:
\begin{enumerate}
    \item \textbf{Small-data AOI pre-training protocol:} We develop an SSL pre-training protocol tailored to small AOI datasets and tiny models under adverse conditions (two-channel input, low variance, black borders, fine structures, and edge compute), and show that MAE outperforms DINO and iBOT in this regime.
    \item \textbf{Efficiency gain:} We demonstrate that MAE pre-training substantially accelerates fine-tuning and yields superior segmentation quality (up to $\sim$50.7\% relative improvement) under a fixed $\sim$8 hour training budget.
    \item \textbf{Practical retrieval-based segmentation:} We introduce image/patch retrieval strategies for AOI achieving training-free, near instant test-time adaptation, and show that patch-level cosine similarity aggregation is often sufficient, while attention-based aggregation is not universally superior for multi-label dense retrieval.
\end{enumerate}

\begin{comment}

Our experiments demonstrate that targeted, single-device retrieval-based segmentation can match or outperform fully fine-tuned segmentation models in low-data regimes and under computational constraints. We further show that simple distance-based $k$-nearest-neighbor retrieval performs on par with more complex attention-based aggregation mechanisms, while avoiding the computational overhead of learned decoders.

Overall, this work presents an efficient and practical alternative to repeated model re-training in AOI pipelines by combining domain-specific self-supervised pre-training with retrieval-based, in-context segmentation. Our results suggest that high-quality semantic segmentation for industrial inspection can be achieved with limited data, reduced training time, and improved robustness to distribution shifts.

\end{comment}
% put contributions  here

\section{Related Work}

\subsection{Self-Supervised Learning for Vision}

Self-supervised learning (SSL) has become a dominant approach for learning transferable visual representations without manual annotation. 
Two families are especially relevant to this work: (i) teacher--student self-distillation methods, exemplified by DINO~\cite{Caron2021DINO}, and (ii) masked image modeling (MIM), exemplified by MAE~\cite{He2022MAE}. 
Hybrid approaches, such as iBOT, combine patch-level prediction with online tokenization, bridging masked prediction and distillation-style training signals~\cite{Zhou2022ibotimagebertpretraining}. 
More recently, large-scale training has produced strong general-purpose encoders, such as \mbox{DINOv2}~\cite{oquab2023dinov2} and DINOv3~\cite{simeoni2025dinov3}, often pre-trained on extremely large and diverse data.

Most SSL progress is demonstrated on generic natural-image benchmarks, whereas industrial domains are typically narrow and data-constrained~\cite{bergmann2019mvtec}. 
In this paper, we focus on \emph{domain-specific} SSL, where pre-training is performed on a comparatively small AOI dataset. We evaluate the effect of different SSL strategies (MAE/DINO/iBOT) under this regime. 
Our emphasis is not on scaling data and model size, but on demonstrating that careful SSL pre-training can be effective and practical for AOI semantic segmentation in small-data settings, and also reduce the labeled fine-tuning budget required to reach acceptable segmentation quality.

\begin{figure*}[t]
    \centering
    \includegraphics[width=0.95\linewidth]{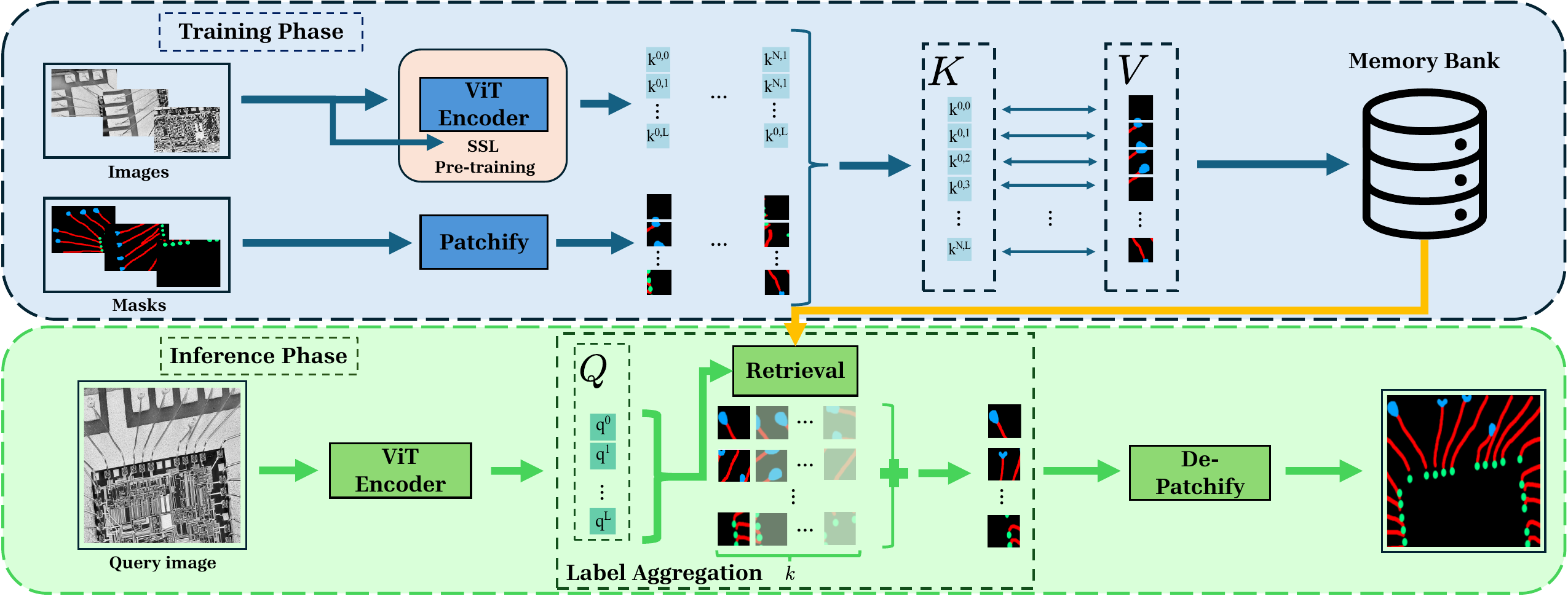}
    \caption{\textbf{Overview of the Retrieval Segmentation Pipeline}. The process follows three stages: (i) a pre-training stage (highlighted in light pink) where a ViT encoder is trained with unlabeled images, (ii) a training phase (light blue region), where training images are encoded using the ViT and stored in the key collection ($K$) in combination with their labels stored in the value collection ($V$), and (iii) an inference phase (light green region) where a query image is encoded into a patch collection $Q$, $k$ nearest patches are retrieved, labels are aggregated and the output mask is re-assembled. The reference device image displayed was adapted from \cite{misterrf2020motorola} under CC~BY-SA~4.0.}
    \label{fig:patch-retrieval}
\end{figure*}

\subsection{Deep Learning for Industrial Inspection and AOI}

Automated visual inspection in semiconductor manufacturing has been widely studied and surveyed~\cite{automate-visual-inspection-survey}, and modern AOI pipelines increasingly incorporate deep learning.
Representative approaches include supervised representation learning for PCB defect classification~\cite{Miao2021, Dai2020}, instance-level detection/segmentation pipelines based on Mask R-CNN for solder joint recognition~\cite{Wu2020}, and cascaded convolutional architectures for SMT solder joint inspection~\cite{Cai2018}. 
Vision transformers \cite{Dosovitskiy2021} have also been explored for industrial visual inspection tasks, indicating that transformer-based models can be effective in industrial inspection~\cite{Hutten2022, CrackFormer}.

For wire-bonded semiconductor inspection specifically, Zhang \etal~\cite{Zhang2024} recently proposed an integrated circuit bonding distance prediction pipeline that leverages modern segmentation architectures as part of a hierarchical inspection pipeline, achieving strong performance in their target setting. 
However, this and many AOI works are typically evaluated under narrow device and acquisition conditions and rely on task- and device-specific engineering~\cite{automate-visual-inspection-survey}, which can limit straightforward reuse when devices or imaging conditions change.
In contrast, we focus on reducing repeated re-training by studying (i) domain-specific self-supervised pre-training in small-data AOI regimes to improve data efficiency, and (ii) retrieval-based, patch-level segmentation that can adapt to new devices by updating a labeled memory rather than re-training a full decoder.

\subsection{Retrieval-Based and In-Context Scene Understanding}

Retrieval-based approaches offer a non-parametric or parameter-efficient alternative to standard decoders by transferring labels from similar examples in a learned embedding space. 
A recent formulation for \emph{in-context} dense prediction is Retrieval-based Scene Understanding (RSU), which first constructs a patch-level memory of embeddings and labels. Then, RSU predicts new masks by aggregating the labels of the retrieved neighbors, via an attention-like label mixing mechanism~\cite{Balazevic2023towards, pariza2025nearfarpatchorderingenhances}. 
This connects naturally to the broader notion of in-context learning~\cite{Brown2020}, where task behavior is conditioned by a context (here, retrieved labeled examples) rather than by gradient-based adaptation.

Despite the strong match between retrieval-based adaptation and AOI needs, retrieval-based patch-level segmentation has seen virtually no adoption in industrial AOI settings. 
Our work adapts these ideas to semantic segmentation of wire-bonded devices and explicitly analyzes design choices that matter in multi-label dense prediction.
\begin{comment}
In particular, we show that naive per-patch label averaging is not optimal for multi-label segmentation retrieval and that simple distance-based $k$NN aggregation can match more complex attention-based aggregation, making the approach attractive for practical deployment.
\end{comment}

\subsection{Architectural Inductive Bias in Vision Transformers}

Vision transformers can be highly effective, but their performance for dense prediction depends strongly on architectural inductive biases and multi-scale feature handling. 
Transformer-based segmentation designs, such as SegFormer~\cite{Xie2021}, emphasize computational efficiency and hierarchical feature construction. 
Similarly, Swin ~\cite{Liu2021swin} and FasterViT ~\cite{Hatamizadeh2023} introduce hierarchical attention with downsampling stages to improve the throughput--accuracy trade-off while preserving multi-scale structure. 
In this work, we leverage such architectures as practical AOI backbones and show that inductive bias (e.g., convolutional stages and downsampling) remains important even after SSL pre-training, particularly for thin and elongated structures, such as wires.

%\subsection{Architectural Inductive Bias in Vision Transformers}

%Vision transformers can be highly effective, but their performance for dense prediction depends strongly on architectural inductive biases and multi-scale feature handling. 
%Transformer-based segmentation designs such as SegFormer emphasize computational efficiency and hierarchical feature construction~\cite{Xie2021}. 
%Similarly, FasterViT introduces hierarchical attention with downsampling stages to improve the throughput--accuracy trade-off while preserving multi-scale structure~\cite{Hatamizadeh2023}. 
%In this work we leverage such architectures as practical AOI backbones and show that inductive bias (e.g., convolutional stages and downsampling) remains important even after SSL pre-training, particularly for thin and elongated structures such as wires.

\begin{comment}

\begin{figure*}[t]
    \centering
    \includegraphics[width=0.8\linewidth]{figures/patch_level_retrieval_diagram_paper_cropped.pdf}
    \caption{Schematic overview of the inference pipeline for our proposed patch-based retrieval segmentation framework. The process follows four stages: (i) encoding of the query image via a feature extractor, (ii) retrieval of the most semantically similar patches from the reference database, (iii) weighted label aggregation, and (iv) spatial re-assembly of the segmented output. The reference database is constructed using a symmetric encoding procedure that indexes embedding-label pairs from the training set.}
    \label{fig:patch-retrieval}
\end{figure*}

\end{comment}

\section{Method}
\label{sec:method}

We address AOI multi-label semantic segmentation for wire-bonded semiconductors under practical constraints: scarce labeled data, low intra-domain variance, two-channel imaging, fine-grained structures (e.g., thin wires), large empty borders, and edge-device compute budgets.
Our approach has three components: (i) domain-specific self-supervised pre-training, (ii) time-budgeted fine-tuning with a lightweight decoder, and (iii) retrieval-based, in-context segmentation that avoids re-training by transferring labels from a labeled memory as detailed in \cref{fig:patch-retrieval}. Moreover, since capturing the fine-grained geometry of the components is of the upmost importance for AOI, we use the mean Intersection over Union (mIoU) metric as our main evaluation tool. 

\subsection{Datasets}
\label{sec:datasets}

Two proprietary datasets of two-channel AOI images of wire-bonded semiconductor devices are used: an unlabeled dataset for self-supervised pre-training and a labeled dataset for supervised fine-tuning and retrieval evaluation. Each channel is a grayscale image captured under a different illumination source, an image set-up relevant for optical calibration and depth estimation in later stages of the inspection process. Furthermore, all images were acquired using the same proprietary optical inspection system, featuring various wire-bonded device models centered on a conveyor belt. 

\textbf{Pre-train dataset (unlabeled).}
The pre-train dataset spans multiple device layouts, bonding configurations, wire geometries (count, thickness, curvature), and packaging material distributions. The resolutions of the images vary substantially (from $\sim$800$\times$800 up to $\sim$8120$\times$8120 pixels), but the vast majority are within the range of 1024$\times$1024 and 2048$\times$2048 pixels. Additionally, most images contain large black borders due to device framing and size. 

The dataset contains $7{,}000+$ unlabeled images collected from $50+$ distinct devices. Its device diversity and wide range of resolutions make it suitable for learning domain-specific representations via self-supervised objectives. This setup is intended to model the realistic scenario where an encoder is pre-trained centrally (e.g., in a data center) and distributed with the AOI system to enable fast downstream adaptation. 
\begin{comment}
To validate the approach we test the performance of the models in higher resolution images on a small test set and find no significant variations in mIoU performance. 
\end{comment}

\begin{figure}[t]
    \centering
    \includegraphics[width=0.65\linewidth]{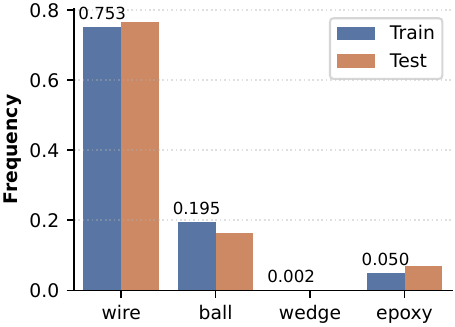}
    \caption{Pixel-wise frequency of the four wire-bond classes in our fine-tune dataset (background not considered). The distribution remains consistent across training (blue) and test (orange) splits, ensuring the model is tested on a representative sample. There is a large class imbalance, notable in the wedge class which consists of less than 1\% of the labeled pixels.} \vspace{-10pt}
    \label{fig:fine-tune-class-distribution}
\end{figure}

\textbf{Fine-tune dataset (labeled).}
The fine-tune dataset contains $625$ labeled images from devices that are also present during pre-training, but under conditions that can induce performance degradation (e.g., covariate shifts from capture and process variation). Image resolutions are within the same range as in the pre-train dataset. The labels are \emph{multi-label} at the pixel level: a pixel may belong to multiple classes, reflecting physically plausible overlaps (e.g., a wire passing over another component). The dataset provides annotations for four foreground classes: \textbf{wire}, \textbf{ball}, \textbf{wedge}, and \textbf{epoxy}. As expected in a semantic segmentation task, the class distribution is strongly imbalanced even without an explicit background class. Figure~\ref{fig:fine-tune-class-distribution} shows the class frequency distribution.

\textbf{Pre-processing.} For training and evaluation, $512\times512$ center crops are extracted to standardize input size. Based on the pre-train dataset statistics, Z-score standardization is also applied in both datasets.

\begin{comment}
\paragraph{Usage in this paper.}
We use the pre-train dataset exclusively for self-supervised learning, and we use the fine-tune dataset for (i) supervised fine-tuning of encoder--decoder segmentation models and (ii) constructing labeled retrieval databases for image-level and patch-level retrieval.
\end{comment}

\subsection{Encoders}
\label{sec:encoders}

Two lightweight vision transformers designed to optimize throughput are utilized: ViT-Tiny, a custom ViT implementation with 6 transformer blocks, a 384-dimensional embedding, and 6 attention heads; and FasterViT-0~\cite{Hatamizadeh2023}, the most lightweight version of this architecture. ViTs generally function by splitting input images into patches, mapping them to a token space, and applying attention, thereby producing patch embeddings. ViT-Tiny additionally includes image level embeddings with a special class token: [CLS]. 

\subsection{Domain-Specific Self-Supervised Pre-training}
\label{sec:pretraining}

We adapt three SSL paradigms to our pre-training dataset: MAE, DINO, and iBOT. In MAE, the SSL objective is to learn meaningful representations via reconstruction. It is done by masking part of the input via the elimination of random patch tokens, which are then reconstructed in pixel space using a light-weight transformer-based decoder head. Learning is driven by the mean squared error (MSE) loss between the unmasked input image and the reconstructed image.

We also leverage view-invariant self-distillation paradigms via DINO and iBOT. These frameworks employ a student-teacher architecture, in which the teacher network is parameterized as an exponential moving average (EMA) of the student’s weights. To mitigate representation collapse, encoders are coupled with a projection head $g$ that maps [CLS] embeddings to a $K$-dimensional probability space, where semantically similar images are clustered together. Specifically, for an input image $I$, a set of augmented views $V_s(I)$ and $V_t(I)$ is generated for the teacher and student, respectively. The student is then trained to match its projected output distribution ($P_s^{(i)}$) with that of the teacher ($P_t^{(i)}$) through a temperature-scaled softmax along all pairs of augmented views between $A_s(I)$ and $A_t(I)$, and along all output dimensions. A running mean $c$ is also subtracted from the teacher’s latent projections to prevent the model from converging to a trivial uniform solution (i.e, centering~\cite{Caron2021DINO}). The resulting self-distillation objective $\mathcal{L}_{\text{DINO}}$ is defined in \cref{eq:dino-loss}.

\begin{align}
P_s^{(i)}(x) &= \frac{\exp(g_s(x)^{(i)} / \tau_s)}{\sum_{j=1}^{K} \exp(g_s(x)^{(j)} / \tau_s)}, \\
P_t^{(i)}(x) &= \frac{\exp((g_t(x)^{(i)} - c) / \tau_t)}{\sum_{j=1}^{K} \exp((g_t(x)^{(j)} - c) / \tau_t)}, \\
\mathcal{L}_{\text{DINO}}(I) &= \sum_{x \in A_t(I)} \sum_{x' \in A_s(I)} \sum_{i=1}^{K} - P_t^{(i)}(x) \log P_s^{(i)}(x')
\label{eq:dino-loss}
\end{align}

While DINO focuses on aligning global [CLS] tokens, iBOT extends this framework by incorporating a patch-level masked image modeling (MIM) objective, where the student is tasked with reconstructing the masked patch-level embeddings using the teacher as a reference.

 \paragraph{Adaptations.} For MAE, a modified pre-training strategy motivated by iBOT and DINO is used, where learnable masked tokens are included in both the encoder and decoder inputs. This is a necessary adaptation to allow the use of the technique in FasterViT, which is not capable of processing an arbitrary number of tokens per-image as required in the original MAE masking strategy~\cite{He2022MAE}.  DINO and iBOT are adapted through hyperparameter tuning and additional data augmentations to ensure convergence on a small AOI-specific dataset. 

\subsection{Semantic Segmentation Fine-tuning}
\label{sec:fine_tuning}

SSL-pre-trained encoders are fine-tuned with a lightweight segmentation head to meet industrial inference time requirements.
We attach an UPerNet~\cite{xiao2018unified} decoder head, which combines multi-scale features with a pyramid pooling module and a feature pyramid network.
For ViT-Tiny, the outputs of the last four transformer blocks are passed through the decoder; for FasterViT-0, its four hierarchical feature maps are used.

\subsection{Retrieval-Based In-Context Segmentation}
\label{sec:retrieval}

To avoid repeated re-training under device changes or distribution shifts, we propose retrieval-based segmentation.
Given a labeled memory bank of reference images, we encode them into an embedding space and transfer labels from nearest neighbors.
Both image-level and patch-level retrieval (summarized in Fig. \ref{fig:patch-retrieval}) are studied.  Since FasterViT yields low-resolution patches within its semantically dense layers, patch-level retrieval is restricted to the ViT-Tiny architecture.
\begin{comment}
patch-level retrieval is the strongest and is our primary retrieval method (summarized in Fig. \ref{fig:patch-retrieval}).
\end{comment}
\paragraph{Key-value Memory Bank Construction}

Let an input crop be patchified into $L$ tokens, and let the encoder output patch embeddings be $Q \in \mathbb{R}^{L \times d}$.

For a reference gallery of $N$ labeled images and embedding dimension $d$, a patch memory is built by a collection of paired normalized embedding keys ($k_j$) and raw patch multi-label value masks ($V_j$):
\begin{comment}
\begin{equation}
K \in \mathbb{R}^{M \times d}, \qquad V \in \mathbb{R}^{M \times (P^2) \times C},
\end{equation}
\end{comment}
\begin{equation}
\mathcal{M} = \left\{ (k_j, V_j) \mid k_j \in \mathbb{R}^d, V_j \in \mathbb{R}^{P^2 \times C} \right\}_{j=1}^M,
\end{equation}
where $M = \sum_{n=1}^N L_n$ is the number of patches stored, $P^2$ is the patch resolution, and $C$ is the number of classes. To maximize IoU, the memory stores the full mask label $V_j$ for patch embedding $k_j$ (instead of average labels as in RSU~\cite{Balazevic2023towards}), an optimization allowed by our relatively small memory bank sizes.

\paragraph{Patch Retrieval and Label Aggregation}

For a query image $Q$, patch embeddings ($q_i$) are computed and normalized, and the top-$k$ neighbors per patch $\mathcal{N}(i)$ are retrieved using cosine similarity with the keys of $\mathcal{M}$. Moreover, the similarity score $s_{i,j}$ is stored for each neighbor $j \in \mathcal{N}(i)$. We then consider the following aggregation mechanisms.
\begin{comment}
For each neighbor $j$ of each input query patch $i$, cosine similarity scores $s_{i,j}$ are then computed:
\begin{equation}
\hat{q}_i = \frac{q_i}{\|q_i\|_2}, \qquad \hat{k}_j = \frac{k_j}{\|k_j\|_2}, \qquad
s_{i,j} = \hat{q}_i^\top \hat{k}_j,
\label{eq:cos_sim}
\end{equation}

and neighbor indices   are obtained via top-$k$ over $s_{i,\cdot}$.
\end{comment}

\textbf{Similarity-based label combination (ours).}
Neighbor labels are directly weighted by normalized similarity ($w^{\text{sim}}_{i,j}$):
\begin{equation}
w^{\text{sim}}_{i,j} = \frac{s_{i,j}}{\sum\limits_{m \in \mathcal{N}(i)} s_{i,m}},
\qquad
\hat{y}_i = \sum_{j \in \mathcal{N}(i)} w^{\text{sim}}_{i,j} \, V_j.
\label{eq:dist_agg}
\end{equation}

\textbf{Attention-based label combination (RSU-style).}
Following an RSU-style aggregation~\cite{Balazevic2023towards}, attention weights ($w^{\text{attn}}_{i,j}$) are computed with temperature $\beta$ and employed to average neighbor labels:
\begin{equation}
w^{\text{attn}}_{i,j} = \mathrm{softmax}_{j \in \mathcal{N}(i)}\!\left(\frac{s_{i,j}}{\beta}\right),
\qquad
\hat{y}_i = \sum_{j \in \mathcal{N}(i)} w^{\text{attn}}_{i,j} \, V_j.
\label{eq:attn_agg}
\end{equation}

The patch predictions $\{\hat{y}_i\}_{i=1}^{L}$ are reshaped back to image resolution, and class-specific thresholds are applied to produce the final multi-label mask.

\paragraph{Image-level Retrieval}

Image-level retrieval functions as a special case of the patch-level algorithm, using global embeddings and full segmentation masks. For a reference gallery of $N$ images with resolution $H\times W$ and dimension $d$, we redefine the query embedding as a single vector $q \in \mathbb{R}^{d}$, and the memory $\mathcal{M}$ defined below, while maintaining the other components unchanged and using only similarity-based label combination:

\begin{comment}
\begin{equation}
Q \in \mathbb{R}^{1 \times d}, \qquad K \in \mathbb{R}^{n \times d}, \qquad V \in \mathbb{R}^{n \times (H \times W) \times C}
\end{equation}
\end{comment}

\begin{equation}
\mathcal{M} = \left\{ (k_j, V_j) \mid k_j \in \mathbb{R}^{d}, V_j \in \mathbb{R}^{(H \times W) \times C} \right\}_{j=1}^N .
\end{equation}

\begin{comment}
\begin{algorithm}[t]
\caption{Patch-level retrieval inference (single image).}
\begin{algorithmic}[1]
\Procedure{PatchRetrieve}{Encoder, PatchDB $(K,V)$, Image, k, Combine, Thresholds, $\beta$}
    \State $C,Q \gets$ Encoder(Image) \Comment{$Q \in \mathbb{R}^{L \times d}$ patch tokens}
    \State Normalize patches: $\hat{q}_i \gets q_i / \|q_i\|_2$
    \State Normalize DB keys: $\hat{k}_j \gets k_j / \|k_j\|_2$
    \State Similarities: $s_{i,j} \gets \hat{q}_i^\top \hat{k}_j$ \Comment{Eq.~\ref{eq:cos_sim}}
    \State Retrieve $\mathcal{N}(i)$ = top-$k$ neighbors of each $i$ by $s_{i,\cdot}$
    \For{$i=1$ to $L$}
        \If{Combine = Attention}
            \State $w^{\text{attn}}_{i,j} \gets \mathrm{softmax}_{j \in \mathcal{N}(i)}\left(\frac{s_{i,j}}{\beta}\right)$ \Comment{Eq.~\ref{eq:attn_agg}}
            \State $\hat{y}_i \gets \sum_{j\in\mathcal{N}(i)} w^{\text{attn}}_{i,j} \, V_j$
        \Else
            \State $w^{\text{dist}}_{i,j} \gets \frac{\exp(s_{i,j})}{\sum_{m\in\mathcal{N}(i)} \exp(s_{i,m})}$ \Comment{Eq.~\ref{eq:dist_agg}}
            \State $\hat{y}_i \gets \sum_{j\in\mathcal{N}(i)} w^{\text{dist}}_{i,j} \, V_j$
        \EndIf
    \EndFor
    \State $\hat{y} \gets$ PatchToImage$(\{\hat{y}_i\}_{i=1}^{L})$
    \For{$c, t_c$ in Enumerate(Thresholds)}
        \State $\hat{y}_{:,:,c} \gets \hat{y}_{:,:,c} > t_c$
    \EndFor
    \State \textbf{return} $\hat{y}$
\EndProcedure
\end{algorithmic}
\label{alg:patch-level-inference}
\end{algorithm}
\end{comment}

\section{Experiments and Results}
\label{sec:experiments}

In this section, we outline our experimental setup and implementation details. Our primary results are then presented alongside the ablation experiments to demonstrate the robustness of our pre-training, fine-tuning and retrieval strategies.

\subsection{Experimental Setup}
\textbf{Data Integrity and Splits.} To ensure a rigorous evaluation and avoid direct leakage, a strict separation between all dataset phases is maintained. Although the labeled and unlabeled SSL datasets overlap in terms of device models, to test proper generalization capacity to unseen images, we ensure that there are no identical overlapping images between the two datasets. The supervised fine-tuning dataset has a 70/30 train-test split, with an additional set of $\sim$100 labeled images for validation. In single-device retrieval experiments, the memory consists of 42 curated images, evaluated on a disjoint set of 24 images from the same device family to ensure that structural priors are not trivialized by image repetition.

\textbf{Baseline Selection and Tuning.} We select ResNet18~\cite{He2016deep}, U-Net++~\cite{Zhou2019unet++} and DeepLabV3+~\cite{Chen2017RethinkingAC} as primary convolutional baselines. These architectures approximate industry standards for high-precision defect detection and high-throughput industrial segmentation, respectively. To ensure a fair comparison, ImageNet-pretrained baselines are trained with an inverted layer-wise learning rate decay; higher learning rates are set in earlier encoder layers to prioritize adaptation of structural filters to the monochrome 2-channel images. 

\textbf{Training Protocol.} All models are trained under a fixed 8-hour computational budget on an NVIDIA RTX 2080 GPU to simulate industrial deployment constraints and hardware conditions. An AdamW optimizer with a weight decay of 0.05 and a cosine learning rate scheduler are utilized.

\begin{comment}
While MobileNetV3~\cite{Howard2019searching} was initially evaluated for its high efficiency, it was excluded from the final results as it failed to reach a functional mIoU threshold for the fine-grained \textit{Wedge} and \textit{Wire} classes.
\end{comment}

\subsection{Pre-training}
\label{sec:pre-training-adaptations}

\textbf{Implementation Details.} For all self-supervised paradigms, a suite of domain-specific data augmentations to reflect AOI-inherent capturing conditions is employed: random vertical/horizontal flips, $90^\circ$ rotations, Gaussian blur ($\sigma \leq 0.5$), contrast jitter, and Gaussian noise ($\sigma^2 \in [0.05, 12.75]$). Training follows a cosine annealing schedule with linear warm up of 100 epochs, adhering to optimization protocols in~\cite{He2022MAE, Caron2021DINO, Zhou2022ibotimagebertpretraining}. We observe that omitting this specific augmentation policy leads to premature representation collapse across all frameworks. 

\textbf{MAE Configuration.} Empirical evaluation indicates that MAE consistently outperforms semantic-centric strategies (DINO, iBOT) in our corpus. We hypothesize that MAE’s pixel-level reconstruction is better suited for dense downstream tasks in domains with low semantic diversity, where cluster-based distillation fails to capture fine-grained geometric features. Therefore, we highlight MAE results and include additional implementation details and evaluations in the supplementary material (\cref{sec:additional-pre-training-details} and \cref{sec:extended_results}, respectively).   

Our MAE adaptation utilizes a decoder consisting of two transformer blocks and a patch-wise regression head, optimized via pixel-wise MSE loss between original and reconstructed masked patches. With a masking ratio of $0.7$, the model effectively captures the underlying data distribution for specialized AOI inspection. Pre-training is performed for a total of 3000 epochs with a batch size of 100 and a base learning rate of $1.5 \times 10^{-4}$ on an NVIDIA A100 GPU.

\subsection{Fine-tuning Optimization, Training Schedule and Loss Function}
AOI fine-tuning should converge quickly and stably.
To this end, a combination of (i) cosine annealing learning rate decay and (ii) layer-wise learning rate decay (LLRD) is used following the fine-tuning protocols established by DINO \cite{Caron2021DINO}.
~\Cref{tab:cosine_lr_ablation}, which displays the performance in the validation set, shows that these choices are crucial, enabling substantial improvements in mIoU. Detailed hyperparameter values can be found in the supplementary material (\cref{sec:fine-tuning-hyperparameters}).

\begin{table}[t]
\centering
\resizebox{0.45\linewidth}{!}{
\begin{tabular}{ccc}
\toprule
\textbf{Cosine} & \textbf{LLRD} & \textbf{mIoU} \\
\textbf{Annealing} \\
\midrule
\checkmark & \checkmark & \textbf{53.7} \\
\checkmark & \xmark & 33.7 \\
\xmark & \checkmark & 19.6 \\
\xmark & \xmark & 22.4 \\
\bottomrule
\end{tabular}
}
\caption{Effect of cosine annealing and layer-wise learning-rate decay (LLRD) on segmentation metrics (best result in \textbf{bold}).} \vspace{-10pt}
\label{tab:cosine_lr_ablation}
\end{table}

\begin{comment}
\begin{table}[ht!]
    \centering
    \renewcommand{\arraystretch}{1.05}
    \resizebox{0.8\linewidth}{!}{
    \begin{tabular}{l|c|c}
    \toprule
       \textbf{Hyperparameter}  & \textbf{ViT-Tiny} & \textbf{FasterViT-0}  \\ \hline
       Learning Rate  &  \(1.0 \times 10^{-3}\) & \(5.0 \times 10^{-4}\) \\ 
       Layer-wise LR Decay & 0.75 & 0.75 \\ 
       AdamW Weight Decay & \(5.0 \times 10^{-2}\) & \(5.0 \times 10^{-2}\) \\
       Warmup Epochs & 5 & 5 \\
       AdamW \(\beta_1, \beta_2\) & 0.9, 0.95 & 0.9, 0.95 \\
       Batch Size & 64 & 32 \\
    \bottomrule
    \end{tabular}}
    \caption{Fine-tuning hyperparameters for transformer-based encoder--decoder segmentation.}
    \label{tab:tranformer-training-parameters}
\end{table}
\end{comment}

\textbf{Loss function.} Because the target of our work is on multi-label classification with overlap, the loss is computed individually for each class and averaged to obtain a single scalar loss value. We experimented with different region-based losses, such as soft DICE~\cite{milletari2016v} and Jaccard~\cite{rahman2016optimizing}, to address the severe class imbalance in our dataset. However, after evaluating several models on the labeled validation set, we determined that the best option for multi-label segmentation, in our use case, is simple class-wise BCE. Detailed results for each loss can be found in \cref{sec:loss-ablation} of the supplementary material.

\subsection{Retrieval Optimization} 
Since our evaluation metric, mIoU, relies on binary masks, a thresholding strategy is defined to convert model outputs into per-class labels. For fine-tuned models (FasterViT, ViT and convolutional baselines), a fixed threshold of $0.5$ is applied. This standardizes the evaluation and avoids optimization bias across the diverse set of architectures tested.

In contrast, retrieval-based techniques produce similarity-driven scores rather than normalized probability distributions. Consequently, performance is highly sensitive to the choice of threshold for each class. Leveraging the negligible training overhead of retrieval, a grid search is performed to obtain the optimal thresholds and number of neighbors ($k$) with 5-fold cross-validation on the supervised training split. The search is performed independently for each class using mIoU as the selection criterion.

To computationally optimize patch retrieval, the Faiss library~\cite{douze2025faiss} is employed to store and perform patch embedding queries on GPU, achieving near-logarithmic inference time in terms of memory bank size. In terms of memory foot-print, storing embeddings and labels for the entire fine-tuning training dataset requires $\sim$~2GB of GPU RAM.      

\subsection{Quantitative Analysis}
\begin{comment}
\textbf{The Impact of Pre-training.} Table~\ref{tab:results-fine-tune-scratch} shows that training from scratch (random initialization) fails to capture fine-grained structures, particularly the \textit{Wedge} class, where IoU scores drop to near-zero ($1.2\%$--$8.2\%$).

\begin{table}[t]
\centering
\caption{\textbf{Ablation of Pre-training Impact.} Comparison of architectures trained from scratch vs. our domain-specific MAE pre-training. Relative improvement (\%) in mIoU is shown in parentheses.}
\label{tab:results-fine-tune-scratch}
\resizebox{\linewidth}{!}{
\begin{tabular}{@{}llccccc@{}}
\toprule
\textbf{Model} & \textbf{Pre-training} & \textbf{mIoU} & \textbf{Epoxy} & \textbf{Wire} & \textbf{Wedge} & \textbf{Ball} \\ \midrule
FasterViT-0 & Scratch & 42.8 & 67.0 & 54.4 & 1.2 & 48.1 \\
FasterViT-0 & \textbf{AOI-MAE} & \textbf{60.3} \scriptsize{(+40.9\%)} & \textbf{79.3} & \textbf{66.7} & \textbf{19.3} & \textbf{75.8} \\ \midrule
ViT-T & Scratch & 35.5 & 64.4 & 33.1 & 2.7 & 41.6 \\
ViT-T & \textbf{AOI-MAE} & 53.5 \scriptsize{(+50.7\%)} & 73.7 & 43.0 & 26.5 & 70.7 \\ \bottomrule
\end{tabular}
}
\end{table}
\end{comment}

\textbf{Comparison to Baseline Supervised Encoders.} In ~\Cref{tab:results-fine-tune}, we compare our best MAE-pretrained configurations against ImageNet-supervised baselines. Our FasterViT-0 + UPerNet achieves the best performance with 60.3\% mIoU, outperforming the baseline ResNet18 + U-Net++ by nearly 8 percentage points. Although fine-tuned models excel in high-support classes, our training-free patch retrieval remains highly competitive (48.1\% mIoU) while offering significantly higher throughput (266.7 crops/s). This trend is also visible in the results for the additional baseline models available in \cref{sec:extended_results} of the supplementary material.

\textbf{Generalization to Foundation Models.} To assess the versatility of our retrieval framework, we evaluated its performance with a frozen ViT-Small DINOv2 encoder pre-trained on the 142M-image LVD142M dataset (RGB images)~\cite{oquab2023dinov2}. To adapt it to two channel input, an artificial channel is added by computing the average of the first two channels. The results demonstrate that the retrieval technique successfully adapts to general-purpose backbones, achieving mIoU scores comparable to the supervised DeepLab baselines. However, this comparison also underscores the significant ``double benefit'' of our proposed pipeline: our custom MAE pre-training provides superior mIoU performance while using less than 1\% of the pre-training data and less than half of the parameters ($\sim$ 10M vs. $\sim$ 22M). Furthermore, our custom backbone achieves $\sim$3$\times$ higher throughput due to its more efficient $16 \times 16$ patch size (DINOv2 uses $14 \times 14$ patches) and smaller size. This suggests that while retrieval is  backbone-agnostic, it reaches its full potential in industrial AOI when paired with domain-specific self-supervision.

\textbf{The Role of Architectural Inductive Bias.} A critical finding in our method comparison is the superior performance of the hybrid FasterViT-0 architecture. While the standard ViT-Tiny achieves a respectable 53.5\% mIoU, FasterViT-0 reaches 60.3\%, outperforming its pure-transformer counterpart. As evidenced by the class scores in ~\cref{tab:results-fine-tune}, this gain is particularly pronounced in the \textit{Wire} class. We attribute this improvement to the integration of techniques such as hierarchical downsampling and overlapping tokenization, which prevent finer details from falling in-between patches. 

\begin{comment}
\begin{table}[t]
\centering
\caption{\textbf{Method Comparison (MAE Pre-trained).} Comparison of our fine-tuned and retrieval methods against ImageNet-supervised baselines.  The relative performance gain from pre-training is displayed between parentheses. Best results are highlighted in bold, while second best are underlined. Throughput is measured on an NVIDIA RTX 2080 GPU.}
\label{tab:results-fine-tune}
\resizebox{\linewidth}{!}{
\begin{tabular}{@{}lccccc@{}}
\toprule
\textbf{Model} & \textbf{Crops/s} & \textbf{mIoU} & \textbf{Wire} & \textbf{Wedge} & \textbf{Ball} \\ \midrule
\textit{Supervised Baselines} & & & & & \\ 
ResNet18 + DeepLab & 87.3 & 43.5 & 50.3 & 10.3 & 46.7 \\
ResNet18 + U-Net++ & 86.9 & 52.4 & \underline{59.1} & 9.2 & 65.8 \\ \midrule
\textit{DINOv2 (LVD142M dataset)} & & & & & \\
ViT-S Patch Retrieval & 90.9 & 43.4 & 48.7 & \textbf{45.8} & 53.4 \\ \midrule
\textit{Ours (MAE Pre-trained)} & & & & & \\
ViT-T + Patch Retrieval & \textbf{266.7} & 48.1 & 47.8 & \underline{44.4} & 61.8 \\
ViT-Tiny + UPerNet & \underline{218.1} & \underline{53.5}\scriptsize{(+50.7\%)} & 43.0 & 26.5 & \underline{70.7} \\
FasterViT-0 + UPerNet & 163.3 & \textbf{60.3}\scriptsize{(+40.9\%)} & \textbf{66.7} & 19.3 & \textbf{75.8} \\ 
\bottomrule

\end{tabular}
}
\end{table}
\end{comment}

\begin{table}[t]
\centering
\caption{\textbf{Method Comparison.} Comparison of our fine-tuned and retrieval methods against ImageNet-supervised baselines. \textit{AOI-MAE} refers to our AOI-specific MAE training. The relative performance gain from pre-training is displayed between parentheses. Best results are highlighted in \textbf{bold}, while second best are underlined. Throughput is measured on an NVIDIA RTX 2080 GPU.}
\label{tab:results-fine-tune}
\resizebox{\linewidth}{!}{
\begin{tabular}{@{}lcccccc@{}}
\toprule
\textbf{Model} & \textbf{Crops/s} & \textbf{mIoU} & \textbf{Epoxy} & \textbf{Wire} & \textbf{Wedge} & \textbf{Ball} \\ \midrule
\textit{Supervised Baselines} & & & & & & \\ 
ResNet18 + DeepLab & 87.3 & 43.5 & 66.7 & 50.3 & 10.3 & 46.7 \\
ResNet18 + U-Net++ & 86.9 & 52.4 & \underline{75.4} & \underline{59.1} & 9.2 & 65.8 \\ \midrule
\textit{Ours (Frozen DINOv2)} & & & & & & \\
ViT-Small Patch Retrieval & 90.9 & 47.8 & 43.4 & 48.7 & \textbf{45.8} & 53.4 \\ \midrule
\textit{Ours (AOI-MAE Pre-trained)} & & & & & & \\
ViT-Tiny + Patch Retrieval & \textbf{266.7} & 48.1 & 38.4 & 47.8 & \underline{44.4} & 61.8 \\
ViT-Tiny + UPerNet & \underline{218.1} & \underline{53.5}\scriptsize{(+50.7\%)} & 73.7 & 43.0 & 26.5 & \underline{70.7} \\
FasterViT-0 + UPerNet & 163.3 &  \textbf{60.3}\scriptsize{(+40.9\%)} &  \textbf{79.3} &\textbf{66.7} & 19.3 & \textbf{75.8} \\ 
\bottomrule

\end{tabular}
}
\end{table}

\textbf{Single-Device Retrieval.} As shown in ~\cref{tab:single-device-results}, when the memory is curated to a uniform device layout, our training-free patch retrieval method reaches 71.5\% mIoU. This massively outperforms the fully fine-tuned ResNet18 + U-Net++ baseline (47.5\%), which struggles heavily in segmenting the wedge class, highlighting retrieval's potential for near-instant model adaptation in production environments.

\begin{table}[t]
\centering
\caption{\textbf{Single-Device Retrieval Performance.} Results of applying patch retrieval with an MAE-pre-trained encoder on a single device layout (best results in \textbf{bold}). We adopt the same hyperparameters used in the main method comparison.} 
\label{tab:single-device-results}
\resizebox{\linewidth}{!}{
\begin{tabular}{@{}lccccc@{}}
\toprule
\textbf{Strategy} & \textbf{mIoU} & \textbf{Epoxy} & \textbf{Wire} & \textbf{Wedge} & \textbf{Ball} \\ \midrule
Patch Retrieval & \textbf{71.5} & \textbf{68.0} & 65.6 & \textbf{78.7} & \textbf{73.5} \\ 
ResNet18 + U-Net++ (Fine-tuned) & 47.5 & 43.3 & \textbf{72.1} & 3.4 & 71.1 \\ \bottomrule

\end{tabular}
}
\vspace{-10pt}
\end{table}

\subsection{Qualitative Analysis and Discussion}

\begin{figure}[t]
    \centering
    \includegraphics[width=\linewidth]{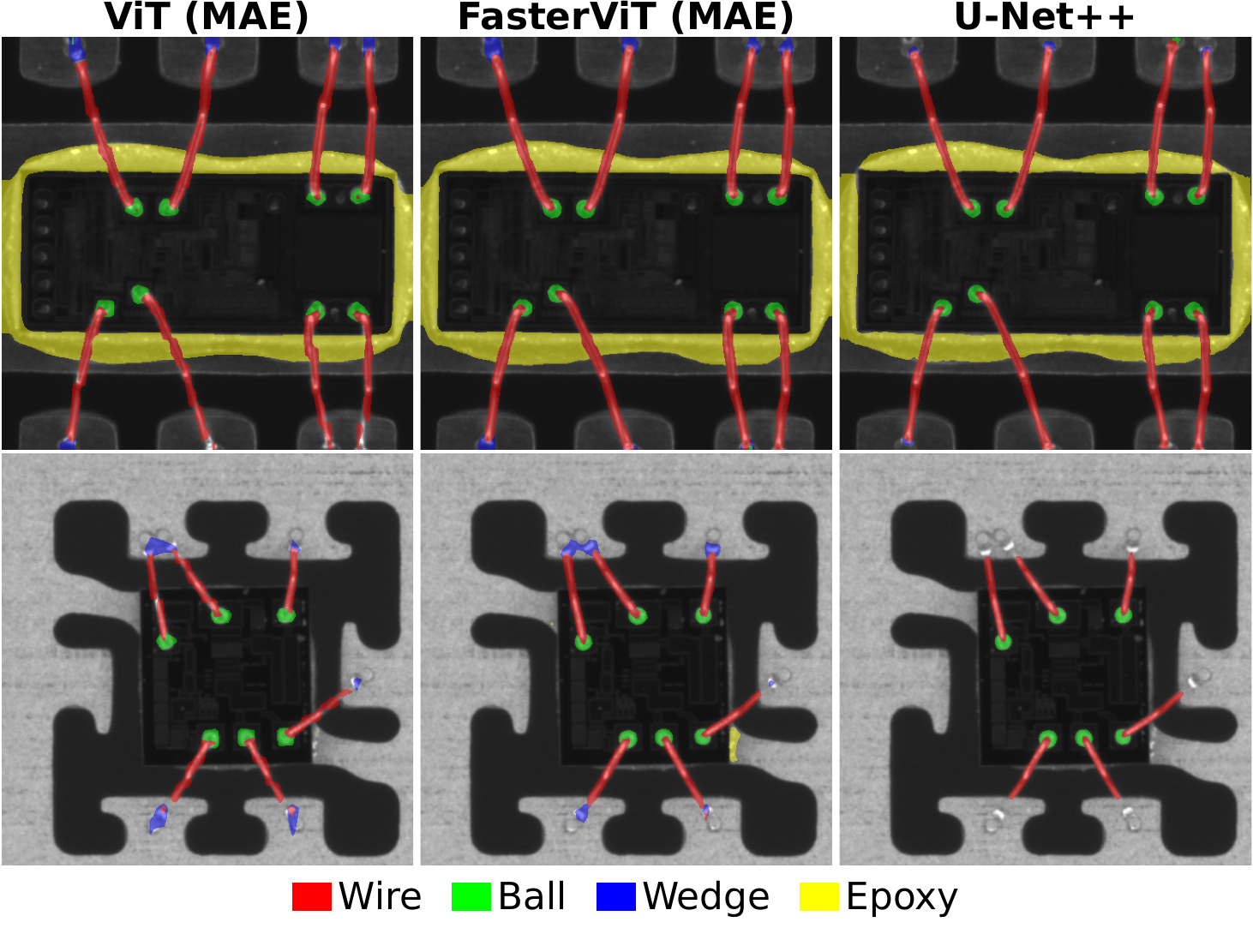}
    \caption{\textbf{Visual Comparison of Fine-tuned Models.} Our MAE-pre-trained, fine-tuned ViT + UPerNet models (left, center) are compared with the ResNet18 + U-Net++ baseline (right). The transformer-based models showcase superior recall for the wedge and epoxy classes.}
    \label{fig:vit-vs-mae} 
\end{figure}

\begin{comment}
\begin{figure}[t]
    \centering
    % Placeholder for your Figure 4: Qualitative Grid
    %\fbox{\rule{0pt}{120pt} \rule{0.9\linewidth}{0pt}}
    \includegraphics[width=1.0\linewidth]{figures/predictions_and_gt_visualization_device_1_paper.pdf}
    \caption{\textbf{Visual comparison of retrieval strategies.} Patch-level retrieval (MAE) showcases superior spatial alignment and retrieval compared to image-level baselines, accurately capturing component boundaries despite layout variations. The images shown are generated using the first channel of representative examples, with the segmentation masks overlaid in multiple colors.}
    \label{fig:qualitative-retrieval}
\end{figure}

\textbf{Spatial Alignment vs. Global Similarity.} A fundamental observation from our qualitative results (Fig.~\ref{fig:qualitative-retrieval}) is the precision gap between image and patch-level retrieval. While image-level retrieval identifies similar device types, it fails to account for sub-pixel positional shifts. Patch-level retrieval aligns predictions with local precision, leading to significantly higher overlap with ground truth, particularly visible in the spatial continuity of wires.
\end{comment}
\textbf{The ``Wedge'' Specialist.} Accurate identification of bonding points is critical in our domain, as they represent a primary failure mode in wire-bonded devices~\cite{packaging-reference-1}. As established previously in \cref{fig:paper-teaser}, our retrieval strategy emerges as a key positive differentiator for the \textit{Wedge} bond class, where standard convolutional decoders typically struggle. This advantage is also present to a lesser degree in our fine-tuned ViTs (see \cref{fig:vit-vs-mae}). In contrast, the convolutional baselines tend to ``over-smooth'' these fine-grained structures, which, in combinations with the large class imbalance, results in poor IoU (3.4\%--9.2\%) due to low recall. By matching similar patches directly, and by pre-conditioning the encoder to the fine-grained structures, our methods achieve 78.7\% IoU on wedge bonds. This suggests that for restricted model scales, in-context retrieval is more effective in preserving fine-grained details than traditional decoder-based adaptation.

\begin{table}[t]
\centering
\vspace{-5pt}
\caption{\textbf{Retrieval Strategy Ablations.} Comparison of different retrieval strategies using 5-fold cross-validation on the fine-tune training split. The best performance of each SSL algorithm is displayed in \textbf{bold}.}
\label{tab:iou-retrieval}
\resizebox{0.7\linewidth}{!}{
\begin{tabular}{@{}lllc@{}}
\toprule
\textbf{Pre-training} & \textbf{Level} & \textbf{Combination} & \textbf{mIoU} \\ \midrule
\textbf{MAE} & Image & Similarity & 26.9 \\
& \textbf{Patch} & \textbf{Similarity} & \textbf{45.7} \\
& Patch & Attention & 44.7 \\ \midrule
DINO & Image & Similarity & 26.6 \\
& \textbf{Patch} & \textbf{Similarity} & \textbf{39.3} \\
& Patch & Attention & 38.3 \\ \midrule
iBOT & Image & Similarity & 29.8 \\
& Patch & Similarity & 33.2 \\
& \textbf{Patch} & \textbf{Attention} & \textbf{40.1} \\ \bottomrule
\end{tabular}
}
\vspace{-10pt}
\end{table}

\textbf{ViT's resolution problem.} As visible in \cref{fig:vit-vs-mae}, a notable finding of our experiments is that standard ViTs often produce ``boxy", low-resolution segmentation masks due to their non-overlapping patch boundaries, a problem less prevalent in FasterViT due to its hierarchical attention and multi-scale features. We expand on other failure modes of our framework in \cref{sec:retrieval-failure-mode} of the supplementary material.

\subsection{Ablations}

\begin{comment}
\subsubsection{Optimization and Regularization}
\label{sec:optimization-ablation}
As shown in Table~\ref{tab:cosine_lr_ablation}, Cosine Annealing is essential for convergence, as static learning rates fail to reach stable minima within our fixed budget. Additionally, LLRD significantly improves performance by preserving the structural feature hierarchy established during SSL pre-training. By moderating the update frequency of earlier layers, LLRD prevents catastrophic forgetting of domain-specific priors while enabling the segmentation head to adapt effectively.
\end{comment}

\paragraph{Retrieval Strategies and Encoders}
\label{sec:retrieval-ablation}
We evaluate the retrieval-based framework across three dimensions: (i) spatial granularity (image-level vs. patch-level), (ii) label aggregation strategy (similarity vs. attention), and (iii) SSL encoder architecture. As summarized in ~\cref{tab:iou-retrieval}, two key trends emerge. First, patch-level retrieval provides a substantial performance gain over image-level baselines, likely due to its ability to handle local deformations and positional shifts common in AOI imagery. Second, similarity-based aggregation consistently shows better performance compared to attention-based retrieval in the MAE and DINO encoders. This suggests that for datasets with low semantic diversity, distance aggregation is sufficient for high-quality retrieval, allowing higher computational efficiency.

\begin{figure}[t]

    \centering

    \includegraphics[width=0.6\linewidth]{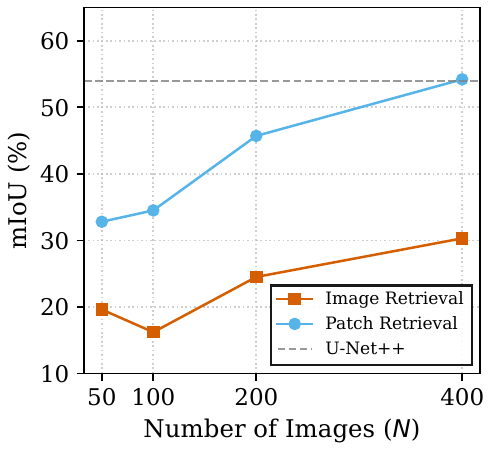}

    \caption{Effect of image gallery size ($N$) in retrieval performance generated through 5-fold cross-validation on the fine-tune training split. The baseline is evaluated on the fine-tune validation split.} \vspace{-10pt}

    \label{fig:retrieval-scalability}

\end{figure}

\paragraph{Retrieval Memory Scalability.} ~\Cref{fig:retrieval-scalability} illustrates the sensitivity of retrieval performance to memory size measured in terms of number of images. We observe that mIoU scales rapidly and near-linearly with the number of labeled images. Crucially, when the memory is expanded to 400 images (33\% less data than used for full fine-tuning), patch-level retrieval exceeds the baseline ResNet18 + U-Net++, demonstrating superior data efficiency.

\section{Limitations}
\label{sec:limitations}

\begin{comment}
While this framework establishes a robust baseline for adaptive semiconductor inspection, several avenues for optimization remain. Our approach focuses on direct SSL pre-training; however, incorporating a knowledge distillation module from a larger encoder could further refine the feature representations. Additionally, while we addressed the extreme class imbalance of the dataset by experimenting with region-bsed losses, further experimentation with specialized losses, such as the Focal Loss~\cite{lin2017focal}, and class weighting may offer additional gains in addressing the class imbalance of bond defects. Furthermore, although our evaluation demonstrates efficacy in $512 \times 512$ center-cropped images, future work is required to assess computational scaling in higher resolution scenarios and  meet the strict segmentation quality requirements of inspection models in production.
\end{comment}
Although our framework establishes a robust baseline for adaptive semiconductor inspection, several avenues for optimization remain. Our approach focuses on direct SSL pre-training; however, incorporating a knowledge distillation module from a larger encoder could further refine the feature representations. Additionally, while the extreme class imbalance of AOI segmentation was addressed through region-based losses, alternative formulations such as the Focal Loss~\cite{lin2017focal} and class re-weighting were considered. Given the substantial overhead of hyperparameter tuning for this setting, we leave such strategies for future work. Lastly, although our evaluation demonstrates efficacy in center-cropped images, future work is required to assess scaling in higher resolution scenarios and meet the segmentation quality requirements of inspection models in production.
\section{Conclusion}
\label{sec:conclusion}

This work introduces AOI-SSL, a data-efficient framework for semiconductor AOI that overcomes data scarcity by combining domain-specific MAE pre-training with a patch-level retrieval mechanism. Our results demonstrate that domain pre-trained ViTs significantly outperform convolutional models and generic ImageNet weights in capturing precise geometric features. Furthermore, the integration of retrieval mechanisms enables near-instant adaptation to new device layouts. Ultimately, this hybrid approach provides a robust solution to the challenges of segmentation in AOI.

% We didn't try distillation from larger pre-trained encoders

% We didn't test other more involved segmentation losses like the focal loss that put more emphasis on hard examples (not the focus of the paper). could be good idea to further push results up

% The dataset remains challending under the required computational efficiency and adaptation time, our results don't yet fully reach industry requirements

% didn't test methods using full resolution, making comparisons not very realistic

\section*{Acknowledgments}
\small {
This work was conducted in collaboration with ASMPT ALSI B.V. within the Vision and AI Laboratory of the Center of Competency (CoC), with additional support from the MARISENS research project. The authors would like to thank the ASMPT engineering team for their technical support, domain expertise, and for providing the datasets used in this study.}
{
    \small
    \bibliographystyle{ieeenat_fullname}
    \bibliography{main}
}
\appendix
\clearpage
\setcounter{page}{1}
\maketitlesupplementary

\begin{comment}
\section{Rationale}
\label{sec:rationale}
% 
Having the supplementary compiled together with the main paper means that:
% 
\begin{itemize}
\item The supplementary can back-reference sections of the main paper, for example, we can refer to \cref{sec:intro};
\item The main paper can forward reference sub-sections within the supplementary explicitly (e.g. referring to a particular experiment); 
\item When submitted to arXiv, the supplementary will already included at the end of the paper.
\end{itemize}
% 
To split the supplementary pages from the main paper, you can use \href{https://support.apple.com/en-ca/guide/preview/prvw11793/mac#:~:text=Delete%20a%20page%20from%20a,or%20choose%20Edit%20%3E%20Delete).}{Preview (on macOS)}, \href{https://www.adobe.com/acrobat/how-to/delete-pages-from-pdf.html#:~:text=Choose%20%E2%80%9CTools%E2%80%9D%20%3E%20%E2%80%9COrganize,or%20pages%20from%20the%20file.}{Adobe Acrobat} (on all OSs), as well as \href{https://superuser.com/questions/517986/is-it-possible-to-delete-some-pages-of-a-pdf-document}{command line tools}.

\end{comment}

\section{Additional pre-training implementation details}
\label{sec:additional-pre-training-details}

\paragraph{DINO}

For the DINO implementation, we attach to our encoders a 3-layer MLP DINO projection head~\cite{Caron2021DINO} with a bottleneck dimension of 2048 and an output embedding dimension of 256. To maintain stability and prevent representation collapse, we employ a teacher EMA decay rate $\tau$ of $0.998$ and a centering momentum $\lambda$ of $0.9$. The student temperature is fixed at $T_s = 0.1$, while the teacher temperature $T_t$ follows a linear warmup from $0.04$ to $0.07$ during the first 30 epochs to encourage diverse feature extraction. These values are based in DINO's original implementation~\cite{Caron2021DINO}. 

An additional element of DINO is its multi-crop strategy, where crops of different sizes of the same images are used to teach scale invariance to the student encoder. Our multi-crop strategy consists of 2 global crops ($226 \times 226$ crops covering an area greater than 50\% of the original image) and 6 local crops ($96 \times 96$ crops covering an area less than 30\% of the original image), with a custom ``black-patch'' filtering logic that re-samples crops if more than 95\% of the pixels are zero-valued, ensuring the model learns from actual structural content rather than dataset artifacts. Given that the color jittering augmentation used in DINO to distinguish teacher and student crops is not applicable to our monochrome images, we use the same augmentations described in \cref{sec:pre-training-adaptations} in both teacher and student views, with the sole difference between them being that the teacher augmentation function $V_t$ returns only global crops, while the student function $V_s$ returns both global and local crops (the optimal number of local crops was determined by an ablation as demonstrated in \cref{tab:dino_num_local_crops}). Moreover, note that we did not pre-train FasterViT with DINO, as it lacks a [CLS] token, and we deemed the training signal obtained only from average pooling of the token embeddings to not be adequate.

\paragraph{iBOT}

The iBOT training protocol extends the DINO training methodology by incorporating a Masked Image Modeling (MIM) loss applied to patch-level tokens, obtained by masking a specific ratio of input tokens to the student. The specific masking strategy used is blockwise masking~\cite{Bao2021beit} with a ratio of 0.3. Following the original iBOT~\cite{Zhou2022ibotimagebertpretraining}, we compute this loss only on the two sets of global teachers and students' crops produced from an input image $I$: $\{x_1^{g_t}, x_2^{g_t}\}$ and $\{x_1^{g_s}, x_2^{g_s}\}$, respectively.  In \cref{eq:ibot-loss} we can see the specific formulation of the loss, where $A_g(I)$ is the collection of pairs of global crops for input image $I$, $u_i$ is the teacher embedding of patch $i$ of the input global crop $x$,  $\hat{u_i}$ is the student embedding of the input global crop $\hat{x}$, $L$ is the number of tokens of input crops $x$ and $\hat{x}$ after being patchified, and $P_\theta^{\text{patch}}$ refers to a shared DINO head that projects the embeddings into a common $K$-dimensional probability space. Moreover, $m_i$ is a scalar value indicating whether patch $i$ is masked ($m_i=1$) or not ($m_i=0$) to ensure that unmasked tokens do not contribute to the loss. 

\begin{align}
A^g(I) &= \left\{ (x, \hat{x}) \mid  x \in \{x_1^{g_t}, x_2^{g_t}\}, \hat{x} \in  \{x_1^{g_s}, x_2^{g_s}\} \right\} \\
\mathcal{L}_{\text{MIM}} (x, \hat{x}) &= \sum_{i=1}^{L} m_i \cdot P_\theta^{\text{patch}}(u_i)^\text{T} \log P_\theta^\text{patch}(\hat{u}_i)\\
\mathcal{L}_{\text{iBOT}} (I) &= \mathcal{L}_{\text{DINO}} (I) + \frac{1}{2} \sum_{x, \hat{x} \in  A^g(I)}\mathcal{L}_{\text{MIM}} (x, \hat{x})
\label{eq:ibot-loss}
\end{align}

The aforementioned shared DINO head is also used to project the image level representations for the DINO loss, in alignment with the original iBOT implementation~\cite{Zhou2022ibotimagebertpretraining}. For the FasterViT-0 backbone, which lacks a native [CLS] token, the DINO loss is calculated on a global average pooled representation of its last feature map. To prevent early-stage divergence in FasterViT, the implementation of this method utilizes a 25-epoch ``warm-up" phase where the MIM loss is removed, allowing the image-level [CLS] representation to stabilize through the DINO objective before introducing patch-level distillation. Both encoder architectures were trained for 3000 epochs using the AdamW optimizer with a base learning rate of $1.5 \times 10^{-4}$, a weight decay of $0.05$, and a batch size of 300 on an A100 GPU.

\paragraph{MAE}

An additional detail regarding our MAE implementation is the protocol used to select the masking ratio. We ablate this hyperparameter to determine the best pre-training configuration in \cref{tab:mae_masking_ablation}. The results demonstrate that high masking ratios (0.7) provide the strongest self-supervised signal for the AOI dataset, while extreme ratios (0.95) lead to a significant drop in representational quality.

\begin{table}[h]
\centering
\caption{Impact of number of local crops on DINO pre-training, evaluated using image-level retrieval (mIoU) on a ViT-Tiny encoder pre-trained on the AOI dataset. Evaluations adopt $k=3$ nearest neighbors and 50 pre-training epochs. The best result is highlighted in \textbf{bold}.}
\resizebox{0.60\linewidth}{!}{
\label{tab:dino_num_local_crops}
\begin{tabular}{@{}cc@{}}
\toprule
\textbf{Number of Local Crops} & \textbf{mIoU (\%)} \\
\midrule
1  & 18.1 \\
3  & 24.0 \\
\textbf{6}  & \textbf{27.4} \\
9  & 26.4 \\
12 & 25.5 \\
\bottomrule
\end{tabular}
}
\end{table}

\begin{table}[h]
\centering
\caption{Effect of masking ratio on MAE pre-training, evaluated using image-level retrieval performance (mIoU) with $k=3$ nearest neighbors (best result in \textbf{bold}). }
\label{tab:mae_masking_ablation}
\resizebox{0.45\linewidth}{!}{
\begin{tabular}{@{}cc@{}}
\toprule
\textbf{Masking Ratio} & \textbf{mIoU (\%)} \\
\midrule
0.20 & 16.8 \\
0.40 & 31.8 \\
\textbf{0.70} & \textbf{32.3} \\
0.95 & 22.0 \\
\bottomrule
\end{tabular}
}
\end{table}

\section{Fine-tuning Hyperparameters}
\label{sec:fine-tuning-hyperparameters}

Table \ref{tab:tranformer-training-parameters} contains the hyperparameters used to fine-tune the transformer based encoders. We fixed the $\beta$ parameters of the AdamW optimizer to the values suggested in DINO~\cite{Caron2021DINO}, and the batch size to maximize GPU  saturation (NVIDIA RTX 2080). The other parameters were experimentally selected to maximize performance in the validation set.   

\begin{table}[ht!]
    \centering
    \renewcommand{\arraystretch}{1.05}
    \caption{Fine-tuning hyperparameters for transformer-based encoder--decoder segmentation.}
    \label{tab:tranformer-training-parameters}
    \resizebox{0.8\linewidth}{!}{
    \begin{tabular}{l|c|c}
    \toprule
       \textbf{Hyperparameter}  & \textbf{ViT-Tiny} & \textbf{FasterViT-0}  \\ \hline
       Learning Rate  &  \(1.0 \times 10^{-3}\) & \(5.0 \times 10^{-4}\) \\ 
       Layer-wise LR Decay & 0.75 & 0.75 \\ 
       AdamW Weight Decay & \(5.0 \times 10^{-2}\) & \(5.0 \times 10^{-2}\) \\
       Warmup Epochs & 5 & 5 \\
       AdamW \(\beta_1, \beta_2\) & 0.9, 0.95 & 0.9, 0.95 \\
       Batch Size & 64 & 32 \\
    \bottomrule
    \end{tabular}}
\end{table}

\begin{comment}
\section{Pre-training Ablations}
\end{comment}
\section{Loss Ablations}
\label{sec:loss-ablation}

The results of the different losses tested are included in \cref{tab:loss_background_ablation}. We additionally ablate the effect of including the background class. The background-less BCE loss consistently outperforms all other losses except in the wedge class. We hypothesize that this dominance of BCE is explained by the additional complexity introduced by the multi-label DICE and Jaccard's optimization objective, which our low-parameter encoders and limited computational budget struggle to solve efficiently. Moreover, we believe that the smaller size and finer details of wedge bonds explain the superior performance of the DICE and Jaccard losses on this class, as they focus on precise mask-prediction overlap rather than simple localization.

\begin{table}[t]
\centering
\caption{\textbf{Ablation on Loss Functions and Background Class.} Impact of various loss formulations and the inclusion of a dedicated background class on segmentation performance. Results were generated using an MAE-pretrained ViT-Tiny encoder paired with a randomly-initialized UPerNet segmentation head. Best results per class are highlighted in \textbf{bold}.}
\label{tab:loss_background_ablation}
\resizebox{\linewidth}{!}{
\begin{tabular}{@{}lcccccc@{}}
\toprule
\textbf{Loss Function} & \textbf{Background} & \textbf{mIoU} & \textbf{Epoxy} & \textbf{Wire} & \textbf{Wedge} & \textbf{Ball} \\
\midrule
\textit{Binary Cross Entropy} & \xmark & \textbf{53.7} & \textbf{66.0} & \textbf{43.7} & 44.3 & \textbf{60.8} \\
& \checkmark & 51.3 & 65.4 & 40.9 & 37.4 & 61.6 \\ 
\addlinespace
\textit{Soft Jaccard} & \xmark & 41.3 & 39.4 & 37.0 & \textbf{51.6} & 37.1 \\
& \checkmark & 35.5 & 37.0 & 34.5 & 44.4 & 26.3 \\ 
\addlinespace
\textit{Soft Dice} & \xmark & 40.9 & 41.0 & 37.7 & 49.9 & 35.1 \\
& \checkmark & 44.0 & 50.3 & 38.0 & 50.5 & 37.1 \\
\bottomrule
\end{tabular}
}
\end{table}

\begin{figure}[t]
    \centering
    \includegraphics[width=1.0\linewidth]{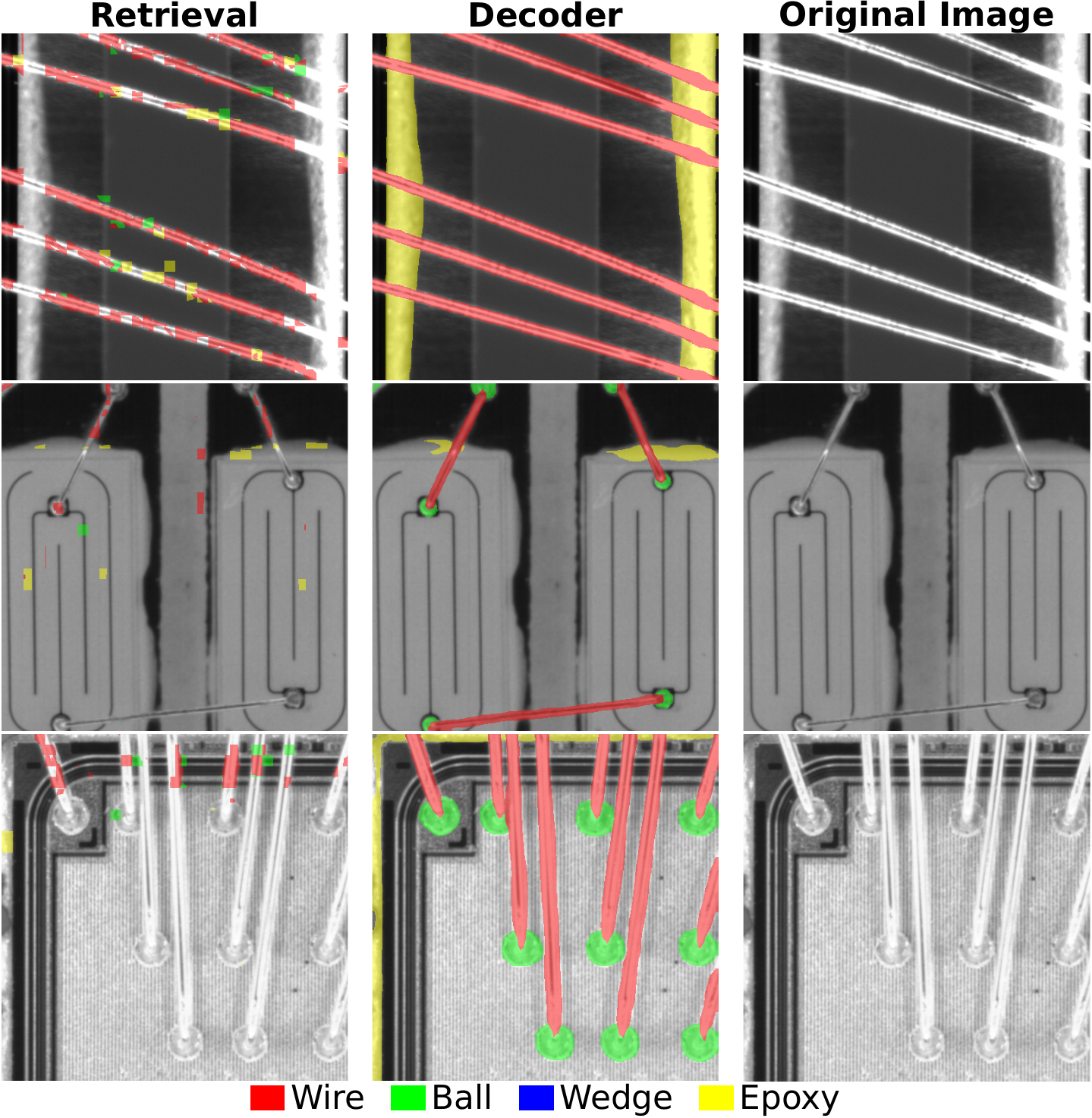}
    \caption{\textbf{Retrieval Failure Cases}. Problematic cases where the patch-level retrieval strategy fails to produce high quality segmentation masks compared to the parametric decoders. Exactly as in \cref{tab:results-fine-tune}, the retrieval strategy used is patch-level retrieval with features produced by an MAE pre-trained ViT-Tiny, while the decoder results employ an MAE pre-trained FasterViT-0 encoder attached to an UPerNet decoder.}
    \label{fig:retrieval-failure-modes}
\end{figure}

\begin{figure}[t]
    \centering
    % Placeholder for your Figure 4: Qualitative Grid
    %\fbox{\rule{0pt}{120pt} \rule{0.9\linewidth}{0pt}}
    \includegraphics[width=1.0\linewidth]{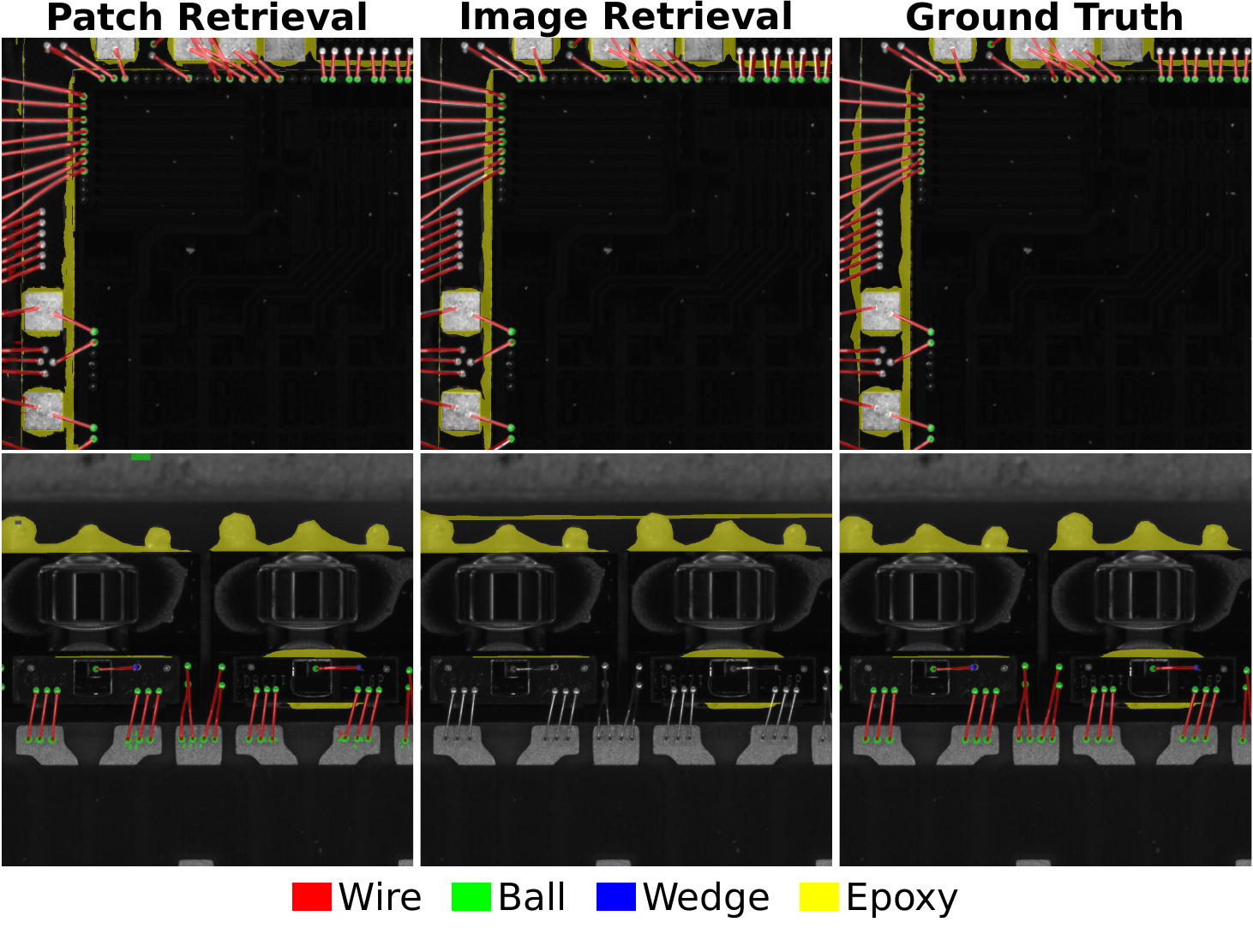}
    \caption{\textbf{Visual Comparison of Retrieval Strategies.} Patch-level retrieval (left) showcases superior spatial alignment and recall compared to image-level baselines (center), accurately capturing component boundaries despite layout variations. These results were generated using an MAE pre-trained ViT encoder as the retrieval backbone.} \vspace{-5pt}
    \label{fig:qualitative-retrieval}
\end{figure}

\begin{table*}[t]
\centering
\caption{\textbf{Comprehensive Method Comparison.} Comparison of our fine-tuned and retrieval methods (MAE, iBOT, and DINO pre-trained) against supervised, ImageNet pre-trained baselines. \textit{AOI Dataset} refers to our custom AOI pre-training dataset. The throughput was measured on an NVIDIA RTX 2080 GPU. The best results are in \textbf{bold}, second best are underlined.}
\label{tab:results-comprehensive}
\resizebox{0.7\linewidth}{!}{
\begin{tabular}{@{}llcccccc@{}}
\toprule
\textbf{Model} & \textbf{Pre-training} & \textbf{Crops/s} & \textbf{mIoU} & \textbf{Epoxy} & \textbf{Wire} & \textbf{Wedge} & \textbf{Ball} \\ \midrule
\textit{Supervised Baselines} & & & & & & & \\ 
MobileNetV3 + DeepLab &  & \textbf{309.8} & 29.9 & 60.6 & 19.3 & 2.8 & 36.8 \\
ResNet18 + DeepLab & ImageNet & 87.3 & 43.5 & 66.7 & 50.3 & 10.3 & 46.7 \\
ResNet18 + U-Net++ &  & 86.9 & 52.4 & \underline{75.4} & 59.1 & 9.2 & 65.8 \\ \midrule
\textit{Ours (Frozen LVD142M DINOv2)} & & & & & & & \\
ViT-S Patch Retrieval & DINOv2 & 90.9 & 47.8 & 43.4 & 48.7 & \textbf{45.8} & 53.4 \\ \midrule
\textit{Ours (AOI Dataset)} & & & & & & & \\
\textbf{FasterViT-0 + UPerNet} & \textbf{MAE} & \multirow{2}{*}{163.3} & \textbf{60.3} & \textbf{79.3} & \textbf{66.7} & 19.3 & \textbf{75.8} \\
FasterViT-0 + UPerNet & iBOT &  & \underline{53.7} & 67.5 & \underline{60.0} & 19.7 & 67.7 \\ \addlinespace
ViT-Tiny + UPerNet & MAE & & 53.5 & 73.7 & 43.0 & 26.5 & \underline{70.7} \\
ViT-Tiny + UPerNet & iBOT & 218.1 & 41.6 & 71.8 & 39.3 & 5.9 & 49.5 \\
ViT-Tiny + UPerNet & DINO &  & 40.0 & 68.7 & 31.6 & 1.3 & 54.2 \\ \addlinespace
ViT-T + Patch Retrieval & MAE &  & 48.1 & 38.4 & 47.8 & \underline{44.4} & 61.8 \\
ViT-T + Patch Retrieval & iBOT & \underline{266.7}  & 45.9 & 46.2 & 43.7 & 41.0 & 52.8 \\
ViT-T + Patch Retrieval & DINO &  & 41.5 & 41.5 & 41.1 & 36.6 & 46.9 \\
\bottomrule
\end{tabular}
}
\end{table*}

\section{Retrieval Failure Modes}
\label{sec:retrieval-failure-mode}

While the proposed retrieval strategies achieve competitive performance without labeled parametric fine-tuning and with significantly higher throughput, they do not yet match the accuracy of the best fine-tuned fully parametric models in the generalized settings of \cref{tab:results-fine-tune}. These performance gaps are particularly pronounced in the \textit{epoxy} and \textit{wire} classes.

\Cref{fig:retrieval-failure-modes} illustrates the primary failure modes underlying these limitations. As shown in the first row, a prominent issue is the lack of spatial continuity on thin, elongated structures. Unlike convolutional decoders, which maintain better local connectivity, our retrieval approach suffers from fragmentation in these regions. We attribute this to the non-overlapping patchification of the ViT-Tiny backbone: when a narrow object does not align perfectly with the patchification grid, individual patches often become dominated by background noise or adjacent classes, diluting the feature representation. Adopting smaller patch sizes or overlapping patches may mitigate this discretization artifact.

Furthermore, the retrieval mechanism exhibits sensitivity to low-contrast images, as demonstrated in the final two rows of \cref{fig:retrieval-failure-modes}. Since the majority of training samples share a consistent contrast profile, extreme outliers reduce the relative similarity and attention scores between the memory bank and the query images. This shift effectively invalidates the class thresholds established during training, resulting in sparse or entirely empty classification masks. To enhance robustness, future work could incorporate contrast-based data augmentations during memory bank construction, ensuring that the reference distribution better accounts for such environmental variations.

A final notable failure mode of our retrieval strategies is their performance within the \textit{epoxy} class in the generalized setting of the main method comparison. As seen in \cref{tab:results-comprehensive}, although most fine-tuned models achieve robust results for this category, retrieval-based strategies exhibit performance degradation. We hypothesize that this stems from the structural characteristics of epoxy, specifically its relatively large size and high geometric irregularity. Effectively capturing such morphological variance likely requires a significantly larger patch memory bank to achieve sufficient representational coverage of the diverse boundary contours. Consequently, this limitation could also benefit from the introduction of augmentations in the construction of the memory bank.

\section{Extended Results}
\label{sec:extended_results}

Detailed results for the exhaustive suite of pre-training and adaptation strategies are provided in \cref{tab:results-comprehensive}. This includes an additional \text{MobileNetV3}~\cite{Howard2019searching} baseline, incorporated to benchmark the throughput--accuracy tradeoff of our proposed retrieval techniques against a faster lightweight architecture. Although MobileNetV3 achieves moderately higher throughput, its mIoU is nearly 50\% \ lower than that of the best performing retrieval strategy, further solidifying the adequacy of retrieval in AOI. These expanded metrics also underscore the significant performance difference between the pre-training regimes mentioned in \cref{sec:pre-training-adaptations}.

 Finally, our expanded qualitative results (Fig.~\ref{fig:qualitative-retrieval}) help to elucidate the significant performance gap between image and patch-level retrieval. While image-level retrieval identifies similar device types successfully, it fails to account for sub-pixel positional shifts. In contrast, patch-level retrieval aligns predictions with local precision, leading to significantly greater overlap with ground truth, particularly visible in the spatial alignment of the wires. Moreover, it also displays higher recall, as it is able to identify individual object patches at fine-grained resolutions without requiring full image-level similarity.


\end{document}